\pdfoutput=1

\documentclass[11pt]{article}

\usepackage[final]{acl}

\usepackage{times}
\usepackage{latexsym}

\usepackage[T1]{fontenc}

\usepackage[utf8]{inputenc}

\usepackage{microtype}

\usepackage{inconsolata}

\usepackage{graphicx}
\usepackage{amsmath}
\usepackage{amssymb}
\usepackage{xcolor}
\usepackage{array}
\usepackage{makecell}
\usepackage{multirow}
\usepackage{colortbl}
\usepackage{subcaption}
\usepackage{tcolorbox}  

\title{Unveiling Confirmation Bias in Chain-of-Thought Reasoning}

\author{
 \textbf{Yue Wan},
 \textbf{Xiaowei Jia},
 \textbf{Xiang Lorraine Li}
\\
\\
 University of Pittsburgh
\\
\texttt{\{yuw253, xianglli\}@pitt.edu}
}

\begin{document}
\maketitle
\begin{abstract}

Chain-of-thought (CoT) prompting has been widely adopted to enhance the reasoning capabilities of large language models (LLMs). 
However, the effectiveness of CoT reasoning is inconsistent across tasks with different reasoning types. This work presents a novel perspective to understand CoT behavior through the lens of \textit{confirmation bias} in cognitive psychology. Specifically, we examine how model internal beliefs, approximated by direct question-answering probabilities, affect both reasoning generation ($Q \to R$) and reasoning-guided answer prediction ($QR \to A$) in CoT. By decomposing CoT into a two-stage process, we conduct a thorough correlation analysis in model beliefs, rationale attributes, and stage-wise performance. Our results provide strong evidence of confirmation bias in LLMs, such that model beliefs not only skew the reasoning process but also influence how rationales are utilized for answer prediction. Furthermore, the interplay between task vulnerability to confirmation bias and the strength of beliefs also provides explanations for CoT effectiveness across reasoning tasks and models. Overall, this study provides a valuable insight for the needs of better prompting strategies that mitigate confirmation bias to enhance reasoning performance. Code is available at \textit{https://github.com/yuewan2/biasedcot}.

\end{abstract}

\section{Introduction}

Chain-of-thought (CoT) prompting \cite{wei2022chain}, which explicitly guides the models to generate intermediate reasoning steps, is one of the most acknowledged prompting strategies for enhancing the reasoning capability of large language models (LLMs). Aside from its benefits of revealing the thinking process in a human-readable format \cite{joshi-etal-2023-machine}, it has proven to be significantly effective in complex reasoning tasks \cite{kojima2022large, zhou2023leasttomost, qi2025mutual}.

To investigate the key factors behind the effectiveness of CoT reasoning, prior studies have examined both the nature of reasoning problems \cite{sprague2024cotcotchainofthoughthelps, feng2023towards, liu2025mind}, the patterns and symbols of the prompts \cite{cotprompteffective}, and the attributes of the CoT rationale \cite{golovneva2023roscoe, prasad-etal-2023-receval}. A key finding across multiple studies is that CoT is particularly useful for symbolic and mathematics reasoning tasks \cite{sprague2024cotcotchainofthoughthelps, feng2023towards}. In contrast, CoT is less effective for non-symbolic reasoning tasks like commonsense reasoning. Moreover, research \cite{liu2025mind} shows that CoT can even hinder performance in tasks where deliberate reasoning negatively impacts human performance. It is also observed that the validity of CoT reasoning contributes only marginally to the CoT performance, whereas query (answer) relevance and reasoning steps ordering play a more important role \cite{towardsunderstandcot}.

\begin{figure}[t]
    \centering
    \includegraphics[width=\columnwidth]{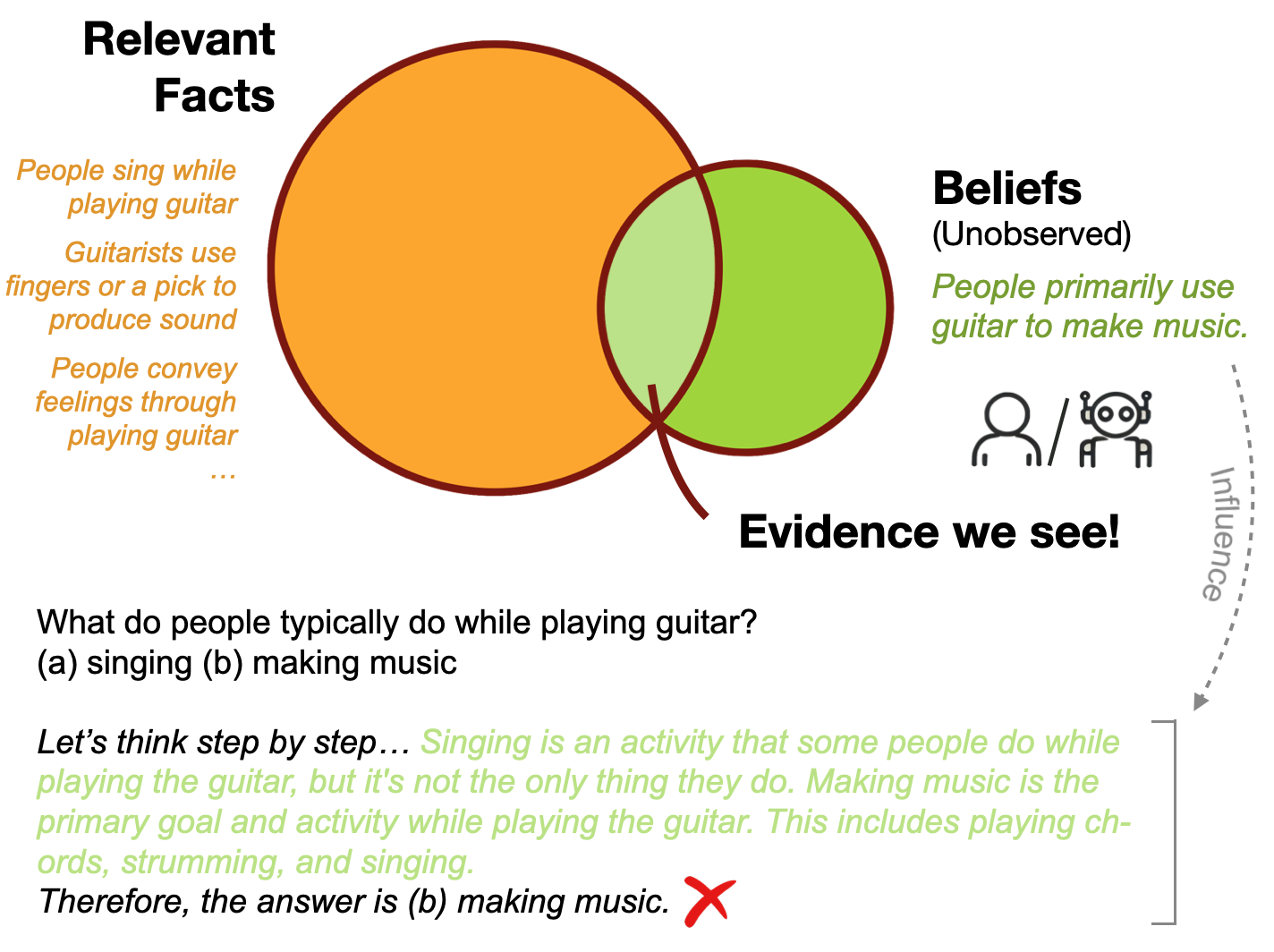}
    \caption{A typical Venn diagram of \underline{confirmation bias} in cognitive psychology, using the example of a commonsensical question. The agent reinforces its internal beliefs and skews its reasoning process towards "making music", while overlooking other relevant facts of playing guitar.
    Notes that the internal belief is unobserved but plays a huge role in decision making.
    }
    \label{fig:illustration}
    \vspace{-15pt}
\end{figure}

In this work, we offer a novel perspective from cognitive psychology to understand the CoT behaviors across reasoning tasks. We argue that, like human beings, LLMs can demonstrate the same patterns of confirmation bias \cite{confirmationbias} that affects the reasoning process. Confirmation bias (Figure~\ref{fig:illustration}) refers to the tendency to selectively retrieve and interpret information in the manner that reinforces preexisting beliefs \cite{confirmationbias}. It is often more pervasive in tasks that require subjective interpretation and prior knowledge compared to those involving formal logic and objective correctness \cite{commonfactorconfirmationbias}. From this perspective, we seek to answer two questions: 1. \textit{How does confirmation bias affect CoT behavior?} and 2. \textit{Why does its influence vary across questions, reasoning types, and LLMs?} We begin by approximating internal beliefs using the direct question-answering probabilities, and the answer confidence as an indicator of beliefs strength. To enable a fine-grained analysis, we decompose CoT reasoning into two stages of reasoning generation ($Q \to R$) and reasoning-guided answer prediction ($QR \to A$). We then perform correlation analysis between beliefs, rationale attributes, and stage-wise performance to explore patterns of confirmation bias across reasoning tasks and LLMs. 



Notably, our experiments reveal patterns of confirmation bias in CoT. The strength of internal beliefs is found to significantly influence CoT performance at both reasoning stages through variations in rationale presentation and how rationale is utilized for answer prediction. 
The extent of CoT improvement also aligns well with the degree to which reasoning tasks are prone to confirmation bias. In addition, we find that "debiasing" internal beliefs becomes even more challenging when they are stronger. This provides a different view of why CoT prompting is most effective in symbolic reasoning tasks (e.g., mathematical reasoning) compared to non-symbolic reasoning tasks, which rely more on contextual and implicit knowledge rather than formal rules for problem-solving. It also sheds light on when CoT can be more reliably trusted. 

In summary, we offer a novel perspective from cognitive psychology in undersanding CoT behavior, showing that patterns of confirmation bias can influence CoT performance across questions, reasoning types, and LLMs. We also propose a new framework for analyzing CoT behavior, which includes the decomposition of the end-to-end accuracy into the performance of $Q \to R$ and $QR \to A$, along with a stratified correlation analysis that connects model internal beliefs with rationale attributes and stage-wise CoT performance. 



\section{Preliminary}
\paragraph{Chain-of-thought} In the conventional chain-of-thought (CoT) \cite{wei2022chain} formulation, a reasoning chain $R$ is explicitly decomposed into intermediate steps $[r_1, r_2, …, r_T]$ given a question $Q$, leading to the final prediction $A$. In convention, each sentence is treated as a reasoning step. Notably, we can factorize CoT into a two-stage process as,
\begin{align*}
    P(A,R|Q)=P(A|Q,R)P(R|Q)
\end{align*}
where the $P(R|Q)$ indicates the reasoning generation stage ($Q \to R$), and $P(A|Q,R)$ corresponds to the stage of reasoning-guided answer prediction ($QR \to A$). The performance of the latter stage is also viewed as the model's faithfulness, which measures the consistency between the predicted answer and the underlying reasoning process \cite{faithfulness1, faithfulness2}. Examining the performance at each stage provides a more fine-grained CoT evaluation.

\paragraph{Confirmation bias} In cognitive psychology, confirmation bias \cite{confirmationbias} is the tendency to seek and interpret information in a way that confirms preexisting beliefs. It is especially pervasive in reasoning processes that rely on subjective interpretation, prior knowledge, and heuristic decision-making \cite{commonfactorconfirmationbias}. In a question-answering setup, beliefs $B$ are often associated with $Q$ and influence the decision as $P(A|Q,B)$. This can be further extended using the CoT formulation:
\begin{align*}
    P(A,R|Q,B)=P(A|Q,R,B)P(R|Q,B)
\end{align*}
which suggests that prior beliefs $B$ may affect both reasoning stages.

\section{Evaluation Methods}
Several challenges exist for exploring confirmation bias in CoT reasoning of LLMs. Firstly, beliefs $B$ are often internal and unobserved. For LLMs, the beliefs associated with a question may come from the prior exposure to question-related content during training, making them hard to measure. 
Second, end-to-end accuracy alone is insufficient for analyzing the effects of $B$ at different stages. A fine-grained correlation analysis requires a stage-wise performance measure, as well as the quantification of $R$'s attributes given $B$.
Third, since we hypothesize that $B$ is a strong prior factor that influences all aspects, it is crucial to develop a method to control its effects in certain analysis. We address each of these challenges in the following sections. We primarily focus on multiple-choice QA questions in this work.


\subsection{Internal Beliefs Quantification}

\paragraph{Direct answer prediction as $B$} 
The actual internal beliefs $B$ are impossible to measure, as they are unobserved and inherently tied to the model's exposure to question-related content during training. However, we argue that the zero-shot answering probability $P(A_i | Q) = \text{softmax} \small {( \frac{1}{T'}  \sum_{t=1}^{T'} \log P(a_{i_t}|a_{i_{1:t-1}}, Q)})$, where $A_i$ denote the $i$th
answer choice given question $Q$ and $T'$ represents the number of tokens in $A_i$, can serve as a proxy. A higher probability indicates that $B$ is more favored towards $A_i$ given $Q$. 

\paragraph{Entropy as strength of $B$}
We then measure the strength of $B$ by the model's confidence over the answer prediction. We leverage the \textit{entropy} of $P(A_i|Q)$ as the measure, where a lower entropy corresponds to higher confidence:
\begin{align*}
    - \frac{1}{C}\sum_{i=1}^n P(A_i | Q) \log P(A_i|Q)
\end{align*}
where $C = \log (n)$ is the normalization factor that scales the entropy between 0 and 1. This normalization enables confidence comparisons across datasets. While entropy is limited to white-box LLMs, we argue that token-level log probabilities provide a direct and clearer reflection of the model’s belief towards the information.

\paragraph{Empirical difficulty as $B$ against $A^*$} To further measure $B$ against the correct answer $A^*$, we compute the log probability difference between $A^\text{*}$ and the highest scored answer choice excluding $A^*$:
\begin{align*}
    \max_{A_i \neq A^*} \log P(A_i | Q) - \log P(A^* | Q)
\end{align*}
We also term this as the \textit{empirical difficulty} of a question. Large negative value means that model is confidently correct about the question (low difficulty), whereas large positive value means the model is confidently incorrect, requiring more efforts to correct $B$ (i.e., greater difficulty). For simplicity, both "entropy" and "empirical difficulty" will only refer to the measures from the direct answering setting in the following sections.



\begin{table*}[]
\centering
\renewcommand{\arraystretch}{1.1} 
\begin{tabular}{lcccccccc}
\Xhline{1.5pt}
\multirow{2}{*}{Datasets} & \multicolumn{2}{c}{Mistral-7B} &  & \multicolumn{2}{c}{Llama3-8B} &  & \multicolumn{2}{c}{OLMo2-7B} \\ \cline{2-3} \cline{5-6} \cline{8-9} 
                          & Direct              & CoT           &  & Direct             & CoT           &  & Direct            & CoT          \\ \hline
CommonsenseQA \cite{commonsenseqa}            & 0.711          &\underline{0.690}         &  & 0.705         & 0.742         &  & 0.623        & 0.766        \\
SocialIQA \cite{social_i_qa}                & 0.651          & \underline{0.653}         &  & 0.564         & 0.631         &  & 0.542        & 0.643        \\
PIQA \cite{piqa}                     & 0.804          & \underline{0.796}         &  & 0.721         & 0.757         &  & 0.666        & 0.713        \\
StrategyQA \cite{strategyqa}                & 0.594          & 0.629         &  & 0.642         & 0.668         &  & 0.572        & 0.607  \\
StrategyQA+F \cite{strategyqa}                & 0.734          & 0.808         &  & 0.760         & 0.817         &  & 0.712        & 0.738  
\\ 
AQuA \cite{aqua}                     & 0.217          & 0.343         &  & 0.291         & 0.480         &  & 0.244        & 0.528        \\ \Xhline{1.5pt}
\end{tabular}
\caption{An overview of chain-of-thought improvement. The underlined scores represent cases where the CoT improvement is either marginal or negative.}
\label{tab:performance}
\vspace{-10pt}
\end{table*}

\subsection{Chain-of-Thought Evaluation}
To analyze the effect of internal beliefs in CoT generation, we evaluate CoT using multiple metrics: (1) \underline{Length} computes the number of tokens in the rationale. (2) \underline{Relevance} \cite{towardsunderstandcot} measures the degree to which the rationale merely explains the question or the predicted answer given the question.
(3) \underline{Explicitness} captures whether at least one reasoning step is explicitly conclusive (e.g., \textit{"... is the most appropriate answer."}). We observe it has a strong influence on subsequent reasoning if presented in the middle steps and the final prediction (Appendix~\ref{sec:explicitness}); (4) \underline{Informativeness}, based on the point-wise mutual information \cite{Bosselut2020DynamicNK, holtzman-etal-2021-surface}, measures how much additional information the rationale provides to improve the CoT prediction; (5) \underline{Sufficiency} evaluates whether the rationale contains enough information to answer the question without the presence of the question. We also include (6) \underline{Relevance}$_\text{Neg}$ and (7) \underline{Explicitness}$_\text{Neg}$, with focuses on how rationale excludes alternative answers. Detailed computations are included in Appendix Table~\ref{tab:metric}. All metrics are hypothesized to correlate with CoT performance.

Since errors can arise at both reasoning stages, it would be insufficient to solely rely on end-to-end performance, Performance$_\text{E2E}$, to conduct the analysis. We thereby extract $A_\text{inter}$ as the \underline{inter}mediate answer supported by the rationale. It is obtained via majority voting from the predicted answers of four advanced LLMs (Appendix~\ref{sec:extract_A_inter}). It is used to evaluate the stage-one beliefs consistency (8) \underline{Consistency}$_{\text{Inter}}=$ {\small $\mathbb{I}\left(\text{argmax}_iP(A_i|Q) = A_\text{inter}\right)$}, and the stage-two performance (9) \underline{Performance}$_{\text{Inter}}=$ {\small $ \mathbb{I}\left(\text{argmax}_iP(A_i|Q,R) = A_\text{inter}\right)$}.




\subsection{Stratified Correlation Analysis}
Based on the quantification of $B$ and the measured attributes of $R$, we perform a correlation analysis
to explore patterns of confirmation bias within CoT. Directly applying correlation analysis to the data has several issues. First, the target factor values may be unevenly distributed, leading to correlation analysis that are biased towards the examples with dominant values. For instance, in our experiments, Mistral-7B \cite{jiang2023mistral7b} has exhibited high confidence (i.e., low entropy) to a large number of questions in CommonsenseQA \cite{commonsenseqa}. Analysis involving entropy may overlook patterns for high entropy questions. 
Second, the question itself is a confounding factor that affects the attributes of $R$, adding noise to the correlation analysis involving $R$.
Third, since we hypothesize that the strength of $B$ (i.e. entropy) may be a dominant factor influencing both $R$'s attributes and performance, directly examining correlations among factors other than entropy could introduce additional confounding effects and lead to a misguided analysis.


To approach these issues, we propose to perform a stratified correlation analysis. Specifically, the factor of interests $\mathbf{z}$ is first discretized into $k$ groups $G$ with equal-width internal $(\mathbf{z}_{max} - \mathbf{z}_{min})/k$. The group assignment is defined as $g(\mathbf{z}_i) = j \text{ if } \mathbf{z}_i \in G_j$. Once the grouping is established, we perform either inter-group or intra-group correlation analysis. 
Inter-group analysis mainly tackles the challenges of imbalanced factor values and data noise. Based on the grouping, factor $\mathbf{x}$ are first aggregated into group-level features: 
\begin{align*}
    \bar{\mathbf{x}_i} = \frac{1}{|S_j|} \sum_{i \in S_j} \mathbf{x}_i
\end{align*}
where $S_j = \{i\:|\:g(\mathbf{z}_i) = j\}$ is the set of indices for observations in group $G_j$. Aggregation essentially ensures that the target factor (e.g., entropy) becomes more uniformly distributed, thereby reducing bias from unbalanced data. Additionally, it helps smooth out the noise originating from individual questions. To avoid overly smoothing the data, we set the number of groups to be sufficiently high, such that the average number of data points within each group is less than 1\%. We then perform correlation analysis with respect to factor $\mathbf{z}$ using the aggregated observations. 

Intra-group analysis focus more on the third challenge. Confounding factor $\mathbf{z}$ is first discretized into $k$ group, and correlation analysis is conducted within each subgroup, considering only questions with similar $\mathbf{z}$ values. 
This allows for a clearer examination of the relationship between key factors, while minimizing the influence of $\mathbf{z}$. It also enables us to further investigate how correlation patterns evolve across different levels of $\mathbf{z}$.


\section{Experimental Setup}

\subsection{Datasets} 

We experiment with five datasets of varying reasoning types: CommonsenseQA \cite{commonsenseqa}, SocialIQA \cite{social_i_qa}, PIQA \cite{piqa}, StrategyQA \cite{strategyqa}, and AQuA \cite{aqua}. We also evaluate StrategyQA+F, where the implicit facts to solve the question are given. Hypothetically, explicitly providing factual knowledge to the models will mitigate confirmation bias from implicit knowledge retrieval, hence leading to larger CoT improvement. 

\subsection{LLMs}   
We choose Mistral-7B \cite{jiang2023mistral7b}, Llama3-8B \cite{llama3}, and OLMo2-7B \cite{olmo2}, three of the most popular and advanced white-box LLMs, for CoT Analysis. 

\subsection{QA Details} 
To compute the direct question-answering prediction, we first apply the softmax function to the average log probability of the answer tokens given the question as $P(A|Q)$. We then select the answer with the highest probability as the prediction. For the CoT prediction, we first generate the rationale from $P(R|Q)$. The zero-shot CoT prompt used in this work is adapted from \citet{cothub} (Appendix~\ref{sec:appendix_cot_prompt}). We then compute $P(A|Q,R)$ in the same manner and extract the CoT prediction. The end-to-end accuracy, denoted as Performance$_\text{E2E}$, measures whether the prediction matches $A^*$. Additionally for CoT evaluation, we measure whether the prediction aligns with $A_\text{Inter}$ (i.e., the \underline{inter}mediate answer extracted from the rationale), regardless of whether it matches $A^*$. This serves as the stage-two accuracy (i.e., Performance$_\text{Inter}$) of the model's ability to faithfully follow the rationale.



\section{Results}
Table~\ref{tab:performance} shows the overall CoT performance. It can be seen that the CoT improvement on non-symbolic reasoning tasks in general falls far behind its improvement on symbolic reasoning problems like AQuA. Mistral-7B even performs worse on CommonsenseQA and PIQA with CoT. This observation aligns well with the findings in \cite{sprague2024cotcotchainofthoughthelps} that CoT primarily improves performance on symbolic and mathematics reasoning tasks. In the following section, we conduct a thorough statistical analysis to understand the performance difference. The following question will be addressed. RQ1. How does confirmation bias affect CoT behavior. RQ2. Why does its influence vary across different questions, reasoning types, and models? 


\begin{figure}[t]
    \centering
    \includegraphics[width=\columnwidth]{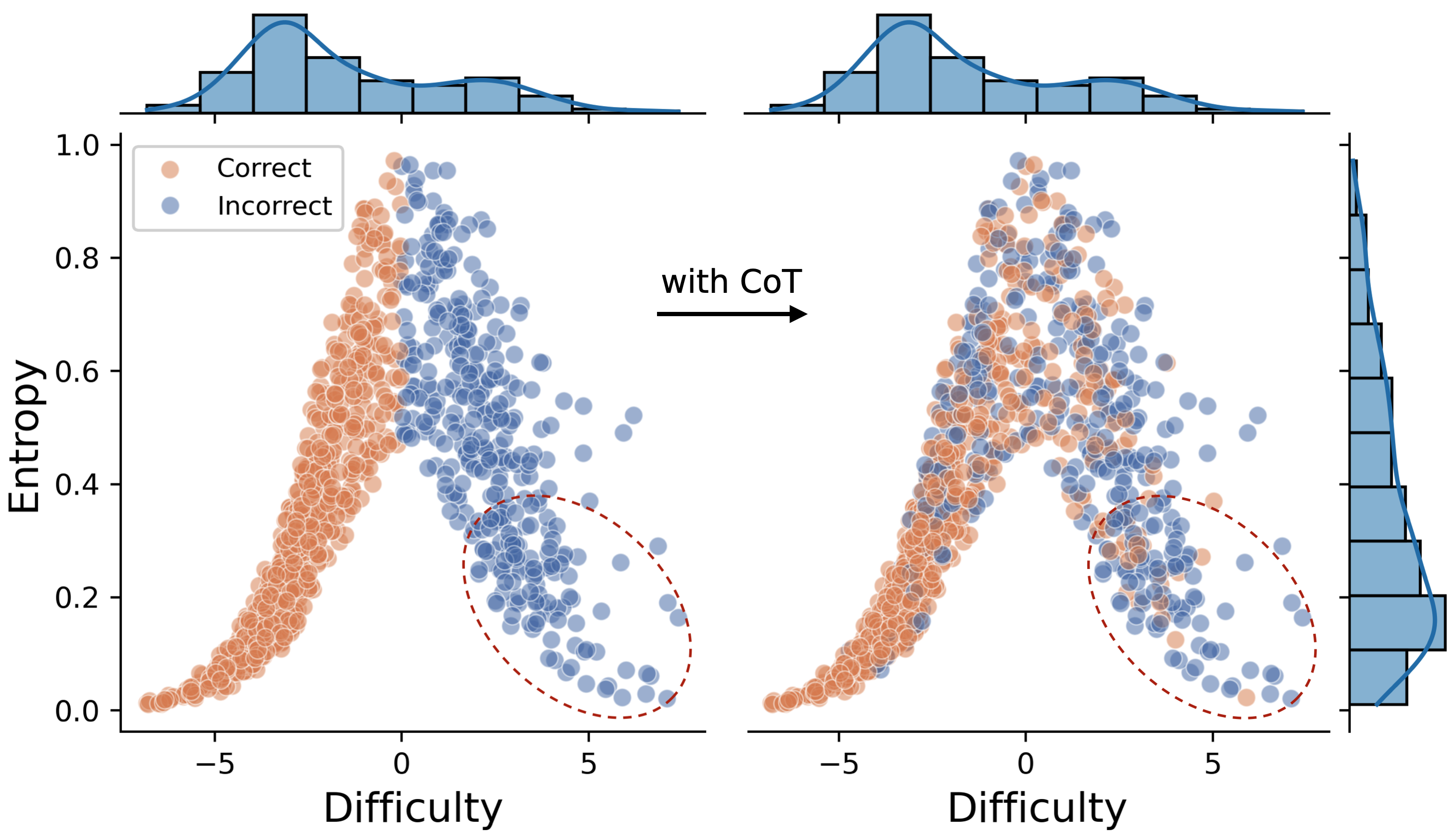}
    \caption{Shift in Performance$_\text{E2E}$ from direct to CoT prediction in relation of entropy and empirical difficulty.}
    \label{fig:cqa_overview}
    \vspace{-8pt}
\end{figure}


\subsection{RQ1: Confirmation bias in $P(A, R|Q, B)$}
To examine internal beliefs in CoT reasoning, we first conduct analysis on the end-to-end CoT performance (Performance$_\text{E2E}$). In this setting, the model is expected to generate both the rationale and answer given the question, which is the typical CoT setup.
We primarily study the CoT behavior of Mistral-7B on CommonsenseQA, which serves as a typical setting for confirmation bias, which we will illustrate in the later section. Additional analyses on other settings are provided in Appendix~\ref{sec:additional_analysis}, which show similar patterns. 

We first visualize the direct Performance$_\text{E2E}$ and CoT Performance$_\text{E2E}$ with respect to Entropy and question Empirical Difficulty in Figure~\ref{fig:cqa_overview}. It is clear to see that questions with stronger beliefs $B$ (lower entropy) are more likely to retain their correctness level regardless of the question difficulty level, suggesting signs of confirmation bias. This partially explains the ineffectiveness of CoT, particularly in regions where the model is confidently wrong initially (as indicated by the red dashed circle). In contrast, questions with weaker beliefs are more prone to fluctuations in predictions. We observe that this behavior arises because questions with weaker beliefs $B$ (higher entropy) are more sensitive to the quality and structure of the generated reasoning, as we will discuss later. 

\begin{figure}[]
    \centering
    \includegraphics[width=\columnwidth]{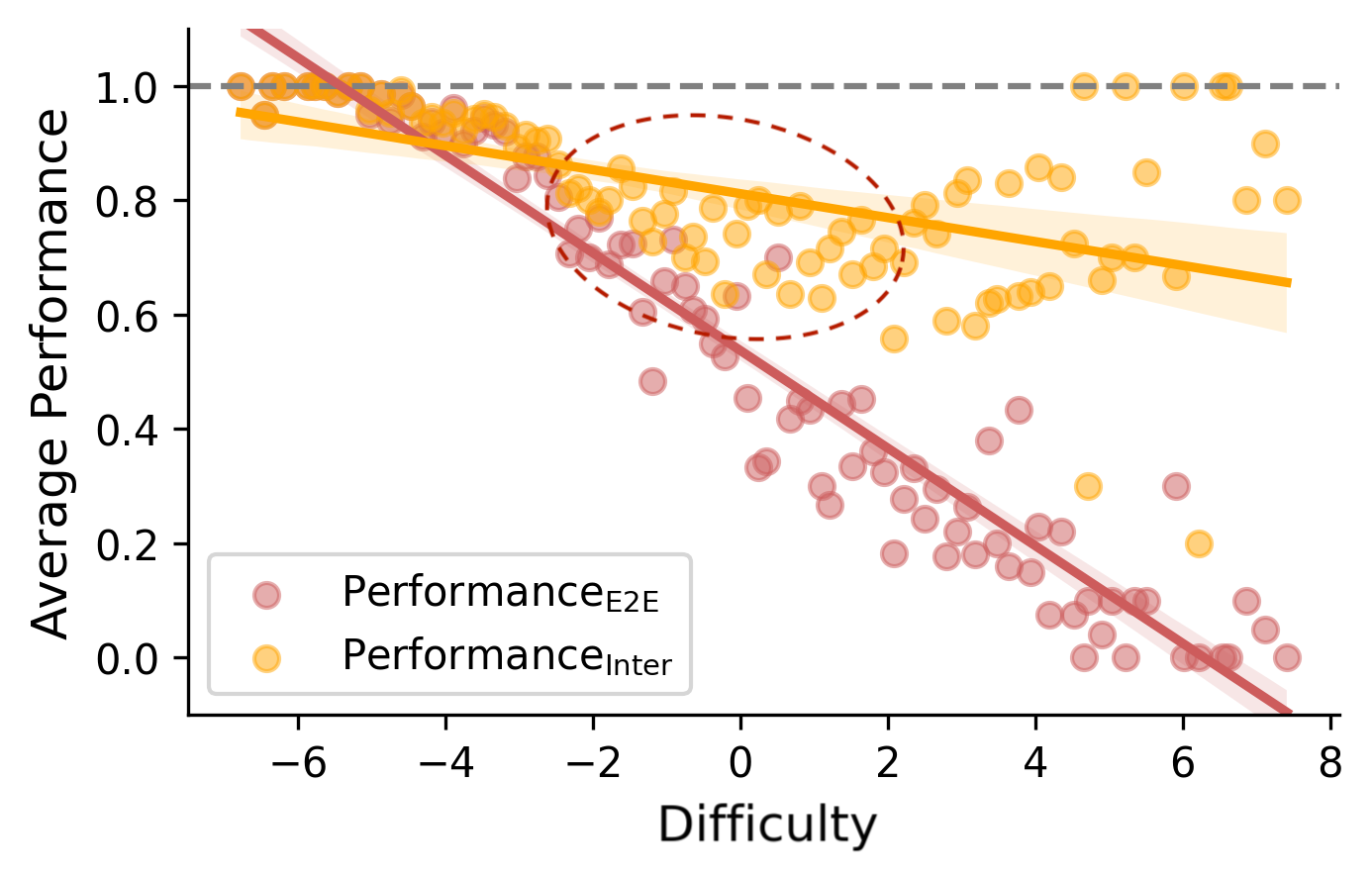}
    \caption{Separation of the Performance$_\text{Inter}$ (i.e., performance of $QR \to A$) from Performance$_\text{E2E}$ (i.e., performance of $Q \to R$ and $QR \to A$) with stratified analysis on empirical difficulty. The grey dashed line represents the perfect performance.}
    \label{fig:seperate_performance}
    \vspace{-8pt}
\end{figure}

We further separate CoT Performance$_\text{Inter}$ from CoT Performance$_\text{E2E}$ and visualize them against the empirical difficulty in Figure~\ref{fig:seperate_performance}. As difficulty increases, Performance$_\text{E2E}$ exhibits a consistent drop, whereas Performance$_\text{Inter}$ remains much more stable. The widening gap between the red and orange lines indicates that errors from the first reasoning stage ($Q \to R$) become more dominant as the model becomes more confidently wrong (i.e., Difficulty$\uparrow$). The gap between the orange and grey (perfect performance) lines reflects the stage-two errors, where the model mis-predicts despite following the "correct" rationale. This is especially true for high entropy questions, as indicated by the red circle.

\begin{figure*}[!ht]
    \centering
    \includegraphics[width=\textwidth]{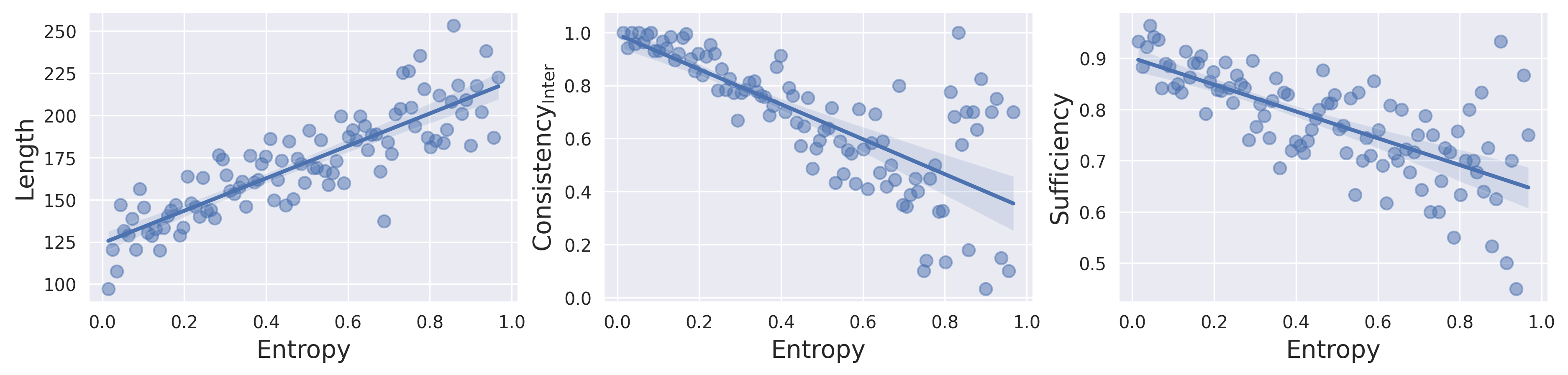}
    \vspace{-20pt}
    \caption{Correlation trends of base entropy (proxy for model's internal beliefs) with CoT Length, Consistency$_\text{Inter}$, and Sufficiency. (Mistral-7B on CommonsenseQA)}
    \vspace{-10pt}
    \label{fig:example_entropy_correlation}
\end{figure*}

\begin{figure}[!ht]
    \centering
    \includegraphics[width=\columnwidth]{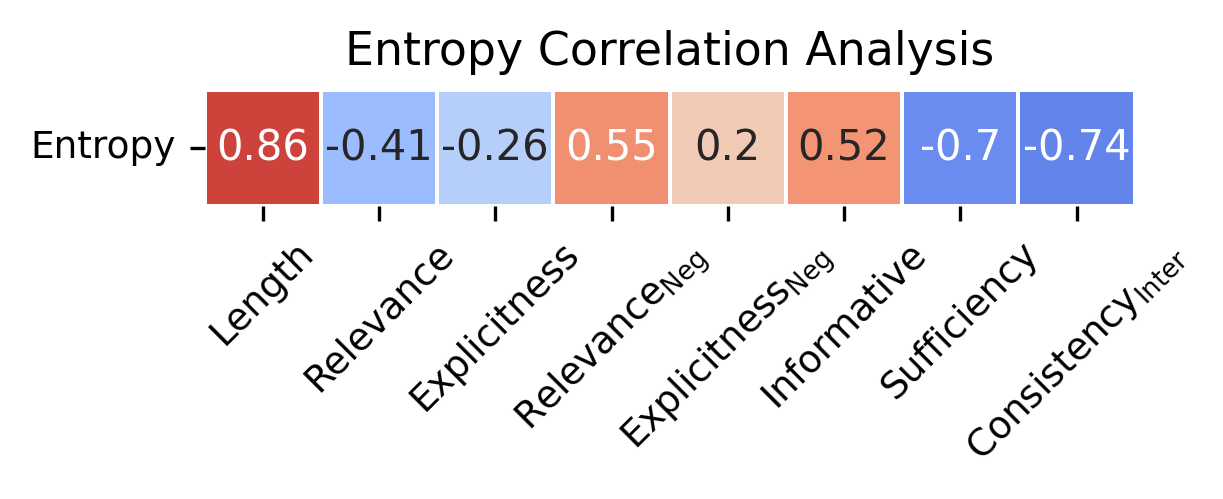}
    \vspace{-20pt}
    \caption{Correlation of Entropy, proxy for strength of model's internal beliefs $B$, with other factors using behaviors of Mistral-7B on CommonsenseQA.}
    \label{fig:entropy_correlation}
    \vspace{-8pt}
\end{figure}

\subsection{RQ1: Confirmation bias disentangled}
To disentangle the impact of confirmation bias, we perform a more detailed analysis of $P(A,R|Q,B)=P(A|Q,R,B)P(R|Q,B)$. Stage 1 analyzes the generated rationale from $P(R|Q,B)$, and stage 2 evaluates the model's performance in faithfully following the generated rationale (Performance$_\text{Inter}$) from $P(A|Q,R,B)$.

\paragraph{Stage 1: $B$ in generated rationale} To investigate how internal beliefs $B$ influence the first stage of $P(R|Q,B)$, we perform the stratified correlation analysis between the entropy values (proxy for the strength of $B$) and $R$'s attributes.
As shown in Figure~\ref{fig:entropy_correlation}, the correlation matrix reveals that the Entropy exhibit strong correlations with six out of eight factors. For questions with strong beliefs (low entropy), models tend to generate shorter reasoning steps, focusing more on explaining the intermediate answer $A_\text{inter}$ (Relevance$\uparrow$) while providing fewer justifications for rejecting alternative choices (Relevance$_\text{Neg}$$\downarrow$). Rationale also tends to be more explicitly conclusive (Explicitness$\uparrow$) for low entropy (strong beliefs $B$) questions, and more likely to explicitly rule out options (Explicitness$_\text{Neg}$$\downarrow$) as $B$ weaken. The negative correlation with Sufficiency may result from the confounding effects of other factors, suggesting that $B$ also affects the overall quality of $R$. 
We also visualize the distribution of the top three correlated attributes in Figure~\ref{fig:example_entropy_correlation}. 


Another key observation is that CoT is more likely to reinforce its original prediction for low entropy questions (Consistency$_\text{Inter}$$\uparrow$). This provides strong evidence of confirmation bias, where prior beliefs affect reasoning outcomes. This may also explain why CoT prompting is more helpful in math reasoning compared to tasks requiring implicit knowledge retrieval \cite{sprague2024cotcotchainofthoughthelps}, as internal belief plays a more significant role in the latter. In order to improve CoT reasoning performance, mitigating the effects of internal belief becomes a crucial problem.

\begin{figure}[!ht]
    \centering
    \begin{subfigure}{\columnwidth}
        \centering
        \includegraphics[width=0.94\linewidth]{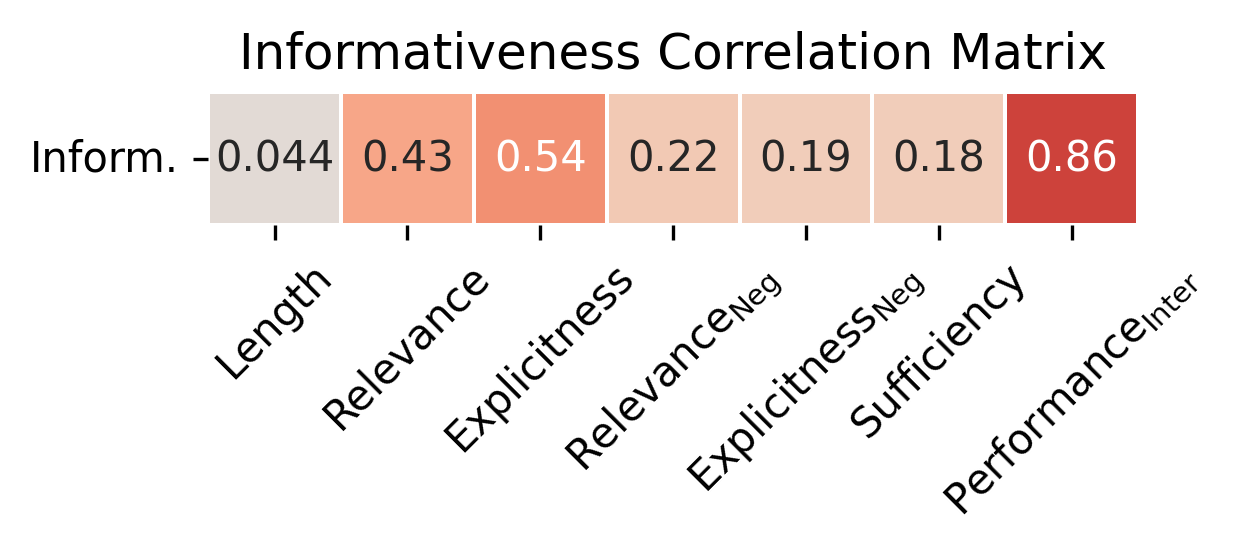}
        \vspace{-5pt}
        \caption{Correlation of Informativeness with other factors.}
        \label{fig:pmi_correlation}
    \end{subfigure}
    \hfill
    \begin{subfigure}{\columnwidth}
        \centering
        \includegraphics[width=\linewidth]{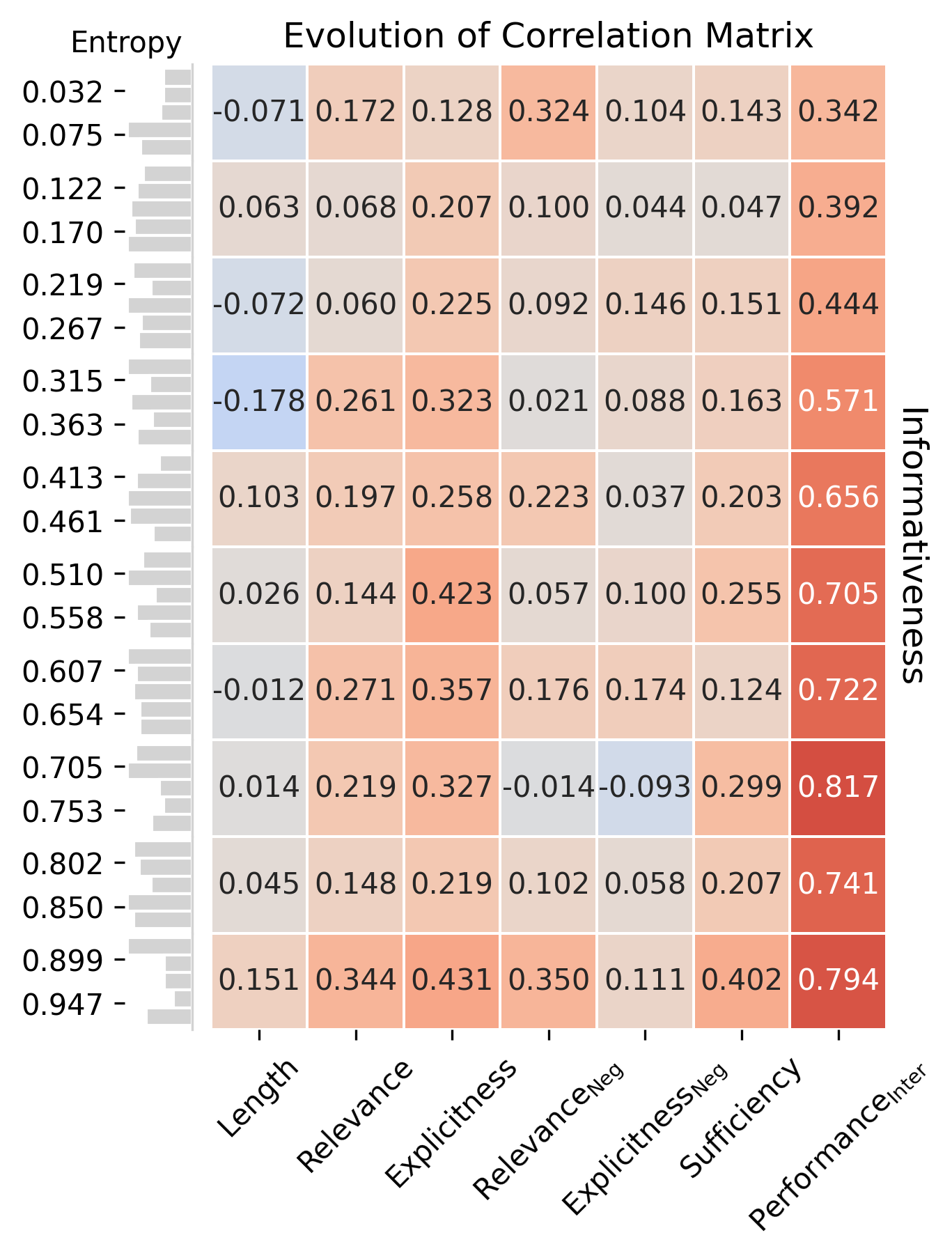}
        \vspace{-20pt}
        \caption{Evolutionary correlation patterns of Informativeness with other factors across different Entropy groups.}
        \label{fig:pmi_correlation_evolve}
    \end{subfigure}
    \caption{Correlation analysis of the role of $B$ in the second reasoning stage of $P(A|Q,R,B)$, using behaviors of Mistral-7B on CommonsenseQA.}
    \label{fig:stage_two_analysis}
    \vspace{-15pt}
\end{figure}

\paragraph{Stage 2: $B$ in rationale-guided answering} 
In this stage, we primarily study the role of $B$ in influencing Performance$_\text{Inter}$. We use Informativeness as the main performance metric for the stratified correlation analysis, as it provides a continuous assessment of models' ability to faithfully following $R$ for predictions. We first examine the general correlation between rationales' attributes and Informativeness. As shown in Figure~\ref{fig:pmi_correlation}, Informativeness appears to be particularly correlated with Relevance and Explicitness on CommonsenseQA by Mistral-7B, which is expected. However, as we already know that entropy (i.e., strength of $B$) also has huge impact on these attributes, we cannot disentangle the effects of $R$ and $B$ in $P(A|Q,R,B)$ from this result.


\begin{figure*}[t]
    \centering
    \includegraphics[width=\textwidth]{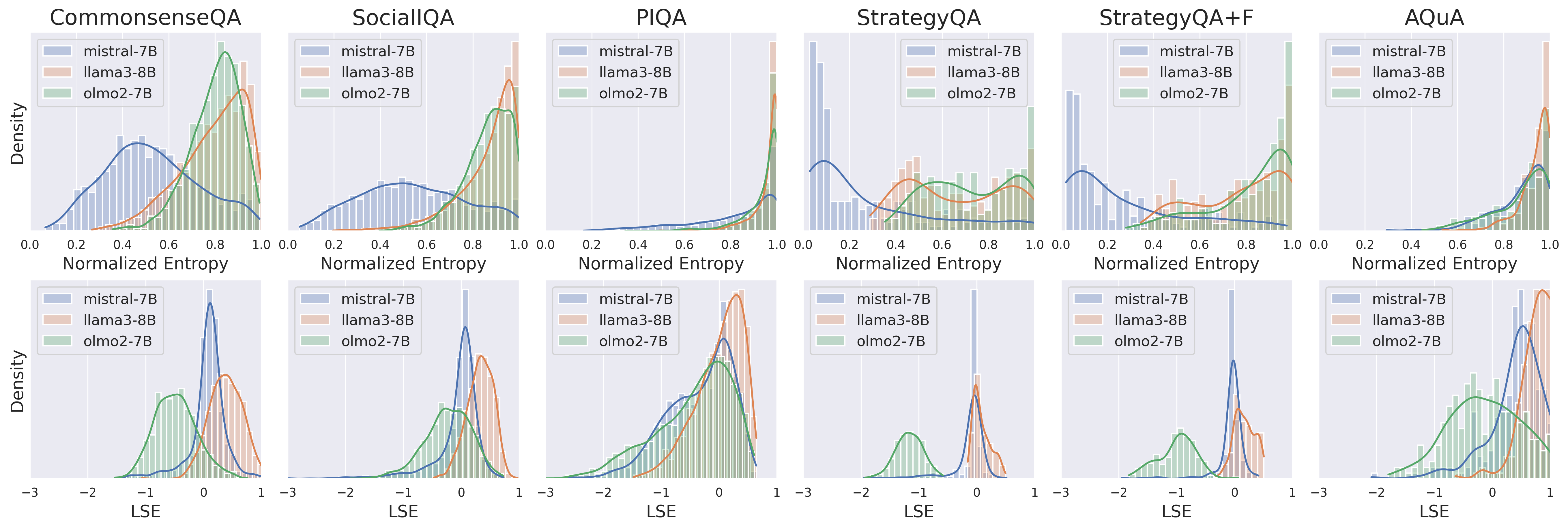}
    \caption{Comprehensive comparison of the question-answering entropy distribution from $P(A_i|Q)$ across the Mistral-7B, Llama3-8B, and OLMo-7B models on six reasoning tasks. Mistral-7B exhibits much lower entropy (stronger beliefs) on large number of questions across nearly all datasets.}
    \label{fig:comprehensive_entropy_histogram}
    \vspace{-8pt}
\end{figure*}

To address this issue, we conduct the intra-group stratified correlation analysis, where the primary grouping is based on Entropy values. For each subgroup, we perform the inter-subgroup analysis on Informativeness. The correlation matrix is shown in Figure~\ref{fig:pmi_correlation_evolve}, where each row represents the correlation between Informativeness and other factors among questions that share similar levels of Entropy. The side column displays the Entropy distribution within each subgroup. One key observation is that the importance of reasoning Relevance, Explicitness, and Sufficiency consistently increases for improved Informativeness as $B$ weaken (questions with higher Entropy). In other words, the model tends to overlook the presentation of the rationale for questions of high confidence, but relying more on its internal beliefs $B$ to infer the answer. The other factors (Length, Relevance$_\text{Neg}$, Explicitness$_\text{Neg}$), on the other hand, do not show clear evolutionary patterns, and are consistently less important. The correlation between Informativeness and Performance$_\text{Inter}$ is lower for low-entropy questions, which results from the cases where high Informativeness is still insufficient to correct an initially confident but incorrect answer. 

\subsection{RQ2: Confirmation Bias Across Settings}

In this section, we provide a comprehensive explanation in why confirmation bias affects CoT performance differently across reasoning types and LLMs. Based on the task subjectivity level and the amount of implicit knowledge required for problem-solving, we rank the datasets based on their vulnerability to confirmation bias as: CommonsenseQA $>$ SocialIQA $\gg$ PIQA $\approx$ StrategyQA $>$ StrategyQA+F $\gg$ AQuA, where the left represents the highest vulnerability (Appendix~\ref{sec:dataset_details}). The CoT improvement of Mistral-7B strictly follows this pattern. In addition, the difference in CoT improvement between StrategyQA and StrategyQA+F further highlights the presence of confirmation bias, such that the removal of potentially biased process of implicit knowledge retrieval leads to greater CoT improvement. Even though the performance of Llama3-8B and OLMo2-7B does not seem to follow the vulnerability hypothesis, this can be explained by the belief differences across models. Since entropy alone cannot distinguish between equally likely and equally unlikely options, we use log-sum-exp ($\text{LSE} =\log \left(\sum_i e^{\log P(A_i|Q)}\right)$) for a finer-grained estimation of beliefs $B$ for cross-model comparison. High entropy with high LSE indicates that the model uncertainty is due to all options are plausible, whereas high entropy with low LSE indicates uncertainty because none of the options are plausible. 

We begin by plotting the Entropy and LSE distribution of the three models against the six reasoning tasks. As shown in Figure~\ref{fig:comprehensive_entropy_histogram}, Mistral-7B demonstrates much lower entropy (stronger $B$) for questions in almost all datasets. In other words, Llama3-8B and OLMo2-7B are inherently less prone to confirmation bias, and are more likely to effectively leverage CoT to improve predictions. This aligns with the correlation results in Figure~\ref{fig:entropy_correlation}, where Entropy and Informativeness are positively correlated. Another observation is that the Entropy distribution of all models shift slightly to the right from StrategyQA to StrategyQA+F, supporting the argument that confirmation bias weakens when implicit knowledge is provided. The reason why OLMo2-7B has marginal CoT improvement on StrategyQA+F can be explained by its LSE distribution. Its overall LSE scale is smaller than that of other models, suggesting that its low confident questions mainly come from equally likely rather than equally unlikely options. This could be another factor  between confirmation bias and CoT behavior that requires further research.


\subsection{Cross-model Debiasing}

Given that different models have different beliefs due to their training processes, another interesting experiment is to evaluate how each model performs using the CoT generated by others. This can be viewed as one model attempting to "debias" the beliefs of another. For convenience, the 
CoT-generating model 
is called the \underline{au}thor, while the one using the CoT for predictions is called the \underline{ex}ecutor. The CoT formulation then becomes $P(A|Q,R_\text{au}, B_\text{ex})P(R_\text{au}|Q,B_\text{au})$. 
If the executor has a different and strong belief ($B_{ex}$) than what the author's rationale supports ($A_\text{inter, au}$), executor's prediction will likely to deviate from $A_\text{inter, au}$, even when $R_\text{au}$ is claimed to be sufficient. 

We first select questions where the zero-shot direct prediction of $P(A_i|Q, B_\text{ex})$ mismatches $A_\text{inter, au}$, and where $R_\text{au}$ is deemed sufficient. We then group these questions into three confidence levels based on the executor's Entropy values and compute the average performance, $\mathbb{I}(\text{argmax}_iP(A_i|Q, R_\text{au}, B_\text{ex}) = A_\text{inter, au})$, for each group. We use Mistral-7B and OLMo2-7B interchangeably as the author and executor, and choose CommonsenseQA and SocialIQA as two datasets that are most vulnerable to confirmation bias.
As shown in Table~\ref{tab:cross_cot}, the executor consistently struggles to follow rationales that contradict its internal beliefs, especially when the beliefs are strong. Even when internal beliefs are weak, the performance still remains suboptimal. This suggests that "debiasing" internal beliefs may be even more challenging than expected. 


\begin{table}[]
\centering
\renewcommand{\arraystretch}{1.2} 
\resizebox{\columnwidth}{!}{
    \begin{tabular}{lcccccc}
    \Xhline{1.5pt}
    \multirow{2}{*}{Dataset} & \multirow{2}{*}{Au} & \multirow{2}{*}{Ex} & & \multicolumn{3}{c}{Performance} \\ \cline{5-7} 
                             &                         &                         & & Strong    & Neural    & Weak    \\ \hline
    \multirow{2}{*}{CQA}     & M                 & O   &                  & 0.5       & 0.636     & 0.776   \\
                             & O                   & M  &                 & 0.510     & 0.718     & 0.833   \\ \hline
    \multirow{2}{*}{SIQA}    & M                 & O   &                  & 0.417     & 0.425     & 0.565   \\
                             & O                   & M  &                 & 0.38      & 0.567     & 0.698   \\ \Xhline{1.5pt}
    \end{tabular}
}
\caption{Performance of \underline{Ex}ecutor using \underline{Au}thor's CoT response (CQA$=$CommonsenseQA, SIQA$=$SocialIQA, M$=$Mistral-7B, O$=$OLMo2-7B).}
\label{tab:cross_cot}
\vspace{-10pt}
\end{table}

\section{Related Works}

\subsection{CoT Prompting}

Chain-of-thought (CoT) prompting \cite{wei2022chain} was introduced to enhance multi-step reasoning in LLMs by explicitly guiding them to generate intermediate reasoning steps, which is proven to be effective in complex reasoning tasks \cite{kojima2022large, nye2022show, zhou2023leasttomost}. Since then, numerous studies have emerged to examine the key factors behind CoT effectiveness. Specifically, researchers \cite{sprague2024cotcotchainofthoughthelps, feng2023towards} found that CoT is particularly useful for symbolic and mathematics reasoning tasks, whereas it only improves marginally on non-symbolic tasks like commonsense reasoning. Liu et al. \cite{liu2025mind} further drew a parallel between CoT and human performance, such that CoT can hinder performance on tasks where deliberate reasoning is counterproductive for humans. Meanwhile, the work in \cite{cotprompteffective} identified consistent patterns and high-quality exemplars in few-shot prompts as two key factors for CoT effectiveness. Several automatic metrics for evaluating reasoning chains were also proposed \cite{golovneva2023roscoe, prasad-etal-2023-receval}. It is observed that CoT performance is influenced more by query relevance and the ordering of reasoning steps, rather than the validity of the reasoning itself \cite{towardsunderstandcot}.

\subsection{LLMs Faithfulness}

Another line of works study the model's faithfulness. It examines how well the model's prediction its true reasoning process, which is also viewed as the model’s self-consistency between the prediction and explanation \cite{faithfulness1, faithfulness2}. This aligns with the stage-2 performance (i.e., Performance$_\text{Inter}$) considered in this work. It is observed that unfaithful behavior of CoT is prevailing across tasks and models, and is related to model sizes \cite{faithfulness2, faithfulness3}, the question-relevant information considered in CoT \cite{faithfulness4}, and the CoT information interacting with the answer \cite{faithfulness4, towardsunderstandcot}. However, these conclusions are only established after CoT rationale is given, while ignoring why CoT is generated differently across questions, tasks and models in the first place. In this work, we also observe that factors like Relevance strongly affect unfaithful behavior in stage-2 prediction. Additionally, our study further highlights the importance of confirmation bias as a strong confounding factor that affects both the CoT generation in the first place and the relationship between rationales’ attributes and model faithfulness.

The work by Bao et al. \cite{bao-etal-2025-likely} is the closet to ours, which studies the CoT consistency and faithfulness by decomposing CoT into explicit problem instruction, CoT generation, and answer prediction. The authors build four types of structural causal model (SCM) among the three parts, and realize that LLMs may have mixed CoT behavior between reasoning and explanation. It further leads to reasoning errors between CoT and predicted answer, and between CoT and the true reason. This is consistent with our findings, and is also closely related to the post-hoc reasoning issue of LLMs. However, our study focuses more on the effects of models’ internal beliefs instead of the explicit instruction given to the model. On the other hand, Bao et al. examines only tasks that rely on formal rules (e.g., mathematical reasoning and logical reasoning), whereas our analytic framework can be applied to a wider range of tasks and models, as long as we can quantify the model confidence before performing any reasoning process.

\subsection{Post-hoc Reasoning}
Researchers also discover that the reasoning process may be post-hoc \cite{faithfulness3, faithfulness5}, where the reasoning process is generated after a conclusion is made. It is observed that post-hoc reasoning is closely related to the unfaithful behavior of CoT. One of the typical assumptions is that the reasoning is more likely to be post-hoc if the models reach the same predictions with and without CoT \cite{faithfulness3, faithfulness5}. Then, CoT reliance can be used as a proxy for unfaithfulness. Our experiments also show a similar but more complex pattern of post-hoc reasoning with CoT reliance. We observe that unfaithfulness alone is more likely to happen for questions with weak beliefs, whereas the prediction consistency with CoT (i.e., inverse of CoT reliance) occurs mostly for questions with strong beliefs. It indicates that post-hoc reasoning may not be directly correlated with CoT unfaithfulness, but again through the confounding effect of confirmation bias.

\section{Conclusion}
In this work, we provide a novel perspective on CoT behavior through the lens of confirmation bias from cognitive psychology. We demonstrate that confirmation bias is pervasive in LLMs, and can substantially impact both reasoning generation and reasoning-guided predictions in the CoT process. In addition, we show that confirmation bias can help explain performance variance across different models and datasets. However, our findings also demonstrate the challenges of "debiasing" confirmation bias, particularly when model beliefs are confidently wrong, underscoring the need for further research.

\section{Limitation} The current work has certain limitations. First, we mainly use the entropy value of zero-shot direct predictions as a proxy for the strength of model beliefs, which limits our analysis to white-box LLMs and multiple-choice questions. A promising extension would be to explore confirmation bias using confidence measures applicable to black-box LLMs and open-ended questions. Hypothetically, open-ended questions could offer a more precise assessment of confirmation bias. The main challenge comes from the need of a precise measurement of the model's confidence over its answer before proceeding any reasoning process for open-ended questions. Unfortunately, uncertainty calibration of LLMs for open-ended generation are much less studied than the multiple-choice QAs. One walkaround is to use LLMs ensembles to generate a pool of candidate answers and perform the analysis as if it were a multiple-choice question \cite{think_twice}. However, the generated pool of answers will introduce additional noise and bias. Since we hypothesize that confirmation bias arises due to LLM memorization, a promising direction for future work is to develop a more appropriate metric that quantifies internal beliefs based on memorization patterns. Second, our experiments only focus on one round of CoT, which overlooks the thought-switching behavior in o1-alike models \cite{openai2024gpt4technicalreport, deepseekai2024deepseekv3technicalreport}. Studying iterative CoT could provide deeper insights into how LLMs revise or reinforce their beliefs.




\bibliography{reference}

\appendix
\renewcommand{\thetable}{S\arabic{table}}
\renewcommand{\thefigure}{S\arabic{figure}}
\setcounter{table}{0}
\setcounter{figure}{0}

\section{Appendix}
\label{sec:appendix}

\begin{table*}[]
\centering
\renewcommand{\arraystretch}{1.2} 
\small
\setcellgapes{2pt} 
\makegapedcells
\begin{tabular}{lccccc}
\Xhline{1.5pt}
Dataset       & Knowledge Type  & Reasoning Type  & Splits & \#Questions & \#Options \\ \hline
CQA \cite{commonsenseqa} & Commonsense     & Commonsense Inference   & validation  & 1221  & 5         \\ \hline
SocialIQA \cite{social_i_qa}     & Social/Cultural & \makecell{Social Inference \\ Theory of Mind \\ Casual Reasoning} & validation & 1954  & 3         \\ \hline
PIQA \cite{piqa}          & Physics        & Casual Reasoning   & validation  & 1838  & 2         \\ \hline
StrategyQA \cite{strategyqa}   & Factual   & Logical Reasoning  & development  & 229   & 2         \\ \hline
StrategyQA+F \cite{strategyqa} & -         & Logical Reasoning  & development  & 229   & 2         \\ \hline
AQuA \cite{aqua}   & Formal    & \makecell{Mathematic Reasoning \\ Logical Reasoning}  & validation & 254   & 5         \\ \Xhline{1.5pt}
\end{tabular}
\caption{Details of the datasets used in this study. "Knowledge Type" indicates the category of knowledge that needs to be implicitly retrieved for solving the task. CQA stands for CommonsenseQA.}
\label{tab:dataset_details}
\end{table*}

\subsection{Datasets Details}
\label{sec:dataset_details}
\paragraph{Statistics} We provide the detailed information of the datasets used in this work in Table~\ref{tab:dataset_details}, including the basic statistics of the datasets used in this work, the knowledge type each dataset focuses on, and the primary reasoning capability required for the task. 

\paragraph{Spectrum of vulnerability to confirmation bias}
On the spectrum of vulnerability to confirmation bias, where the left represents the highest vulnerability, we argue that the approximate ordering of the datasets is: CommonsenseQA $>$ SocialIQA $\gg$ PIQA $\approx$ StrategyQA $>$ StrategyQA+F $\gg$ AQuA. For starters, confirmation bias is more influential in tasks that required subjective interpretation rather than objective inference \cite{commonfactorconfirmationbias}. This makes AQuA the least susceptible to confirmation bias, as it relies on formal logic and structured systems to solve the problems. In addition, mathematical reasoning problems typically have a single correct answer, leaving little room for confirmation bias to distort the reasoning process. StrategyQA and PIQA depend on factual and physical knowledge, making them more objective than subjective. However, confirmation bias can still influence how knowledge is implicitly and selectively retrieved, making both datasets more susceptible to confirmation bias compared to AQuA. On the other hand, StrategyQA+F, where the implicit knowledge required for solving StrategyQA is explicitly provided, is reduced to a pure logical reasoning problem. In contrast, both CommonsenseQA and SocialIQA rely on implicit and subjective understanding of everyday commonsense knowledge, social norms, and cultural conventions, making them the most vulnerable to confirmation bias. Moreover, commonsense reasoning problems may often involve multiple reasoning pathways, where different perspectives can lead to different yet plausible conclusions \cite{cheng-etal-2024-every}. This further increases the susceptibility to confirmation bias. CommonsenseQA is slightly more affected than SocialIQA due to the way we approximate the strength of internal beliefs $B$. Since we use entropy to measure the confidence or strength of $B$, the computation becomes more reliable when more answer options are available.

\subsection{Implementation Details}

All experiments in this work are conducted using the Huggingface framework \cite{wolf2020huggingfacestransformersstateoftheartnatural}. Specifically, we use the \textit{mistralai/Mistral-7B-Instruct-v0.2} snapshot for Mistral-7B, \textit{meta-llama/Meta-Llama-3-8B-Instruct} for Llama3-8B, and \textit{allenai/OLMo-2-1124-7B-Instruct} for OLMo2-7B. We use greedy decoding to generate the rationale used for the performance Table~\ref{tab:performance}. Meanwhile, we use nucleus sampling to generate 10 different CoT responses for the analysis of confirmation bias. For nucleus sampling, both \texttt{temperature} and \texttt{top\_p} values are set to 0.9. We use the \textit{roberta-large-mnli} snapshot for the entailment model used for CoT evaluation (Table~\ref{tab:metric}).

\subsubsection{Chain-of-thought Prompts}
\label{sec:appendix_cot_prompt}
The zero-shot chain-of-thought prompt used in this work is modified from the work in \cite{cothub}:

\begin{tcolorbox}[colback=gray!5, 
    colframe=gray!50, 
    fontupper=\ttfamily\footnotesize, 
    boxrule=0.75pt,
    left=5pt,
    right=5pt
]
You will be given a question at the end, for which you are to select the most appropriate answer by indicating the associated letter. Please first output step-by-step reasoning about how to solve the question. Then, in the last sentence, output which answer is correct in the format of "Therefore, the answer is ...".
\newline
\newline
Question: <question>
\newline
Answer choices: (a) <choice a> (b) <choice b> (c) <choice c> ...
\newline
\newline
Let’s think step by step. To solve the question, we need to 
\end{tcolorbox}

Even though models are instructed to predict the answer in the given format, the generated results may still deviate from it, making it challenging to extract the prediction precisely. Therefore, to better measure $P(A|Q,R)$, we remove the last conclusive sentence from $R$ and compute the answering probability by applying the softmax function to the average log probability of the answer tokens.

\subsubsection{Extraction of Intermediate Answer}
\label{sec:extract_A_inter}
Since errors can occur in the second reasoning stage of $QR \to A$, we extract $A_\text{inter}$ as the \underline{inter}mediate answer choice supported by the reasoning process and measure both stage-one Consistency$_\text{Inter}$ and stage-two Performance$_\text{Inter}$. The extraction is performed by prompting advanced LLMs to select answer based on the question and the generated CoT. In this work, we leverage four advanced LLMs with majority voting to extract $A_\text{inter}$: 1. GPT-4o-mini \cite{openai2024gpt4technicalreport} 2. Llama-3.3-70b-instruct \cite{llama3} 3. Claude-3.5-Sonnet, and 4. DeepSeek-V3 \cite{deepseekai2024deepseekv3technicalreport}. We use the OpenRouter platform \cite{openrouter2025} to access these LLMs.  
Since most of these models are black-box LLMs, we prompt the models to output answers directly with additional instructions shown below. Even though these models can still make mistakes, we believe their advanced reasoning capabilities, combined with the majority voting protocol, can minimize errors at best.  

\begin{tcolorbox}[
    colback=gray!5, 
    colframe=gray!50, 
    fontupper=\ttfamily\footnotesize, 
    boxrule=0.75pt,
    left=5pt,
    right=5pt
]
Question: <question>
\newline
Answer choices: (a) <choice a> (b) <choice b> (c) <choice c> …
\newline
Rationale: <generated chain-of-thought reasoning>
\newline
\newline
Select the most appropriate answer that can be concluded from the given rationale. You must choose only ONE answer. Directly output in the format of "Therefore, the answer is ...".'
\end{tcolorbox}

\subsection{Computation Budget}
The total computation time for CoT experiments, including both CoT generation and CoT evaluation, takes about 200 computation hours on a single A100 GPU. 

\subsection{Explicitness versus Performance}
\label{sec:explicitness}
We observe that rationale explicitness is key factor in the model's ability to follow the reasoning path $P(A|Q,R)$. We first group the questions based on their Explicitness and Explicitness$_\text{Neg}$ levels, and compare their average stage-two performance (Performance$_\text{Inter}$). We evaluate performance under three settings: Mistral-7B on CommonsenseQA and SocialIQA, and OLMo2-7B on CommonsenseQA. As shown in Table~\ref{tab:explicit}, questions in general yield higher performance when at least one of the reasoning steps is explicitly conclusive. On the other hand, being explicit towards why the alternative options are wrong (Explicitness$_\text{Neg}$) shows mixed patterns. This can be explained by LLMs' difficulty in applying the process of elimination \cite{noteasybeingwrong}.

\subsection{Informativeness versus Performance}
\label{sec:informativeness}
\begin{figure}[!t]
    \centering
    \includegraphics[width=0.95\columnwidth]{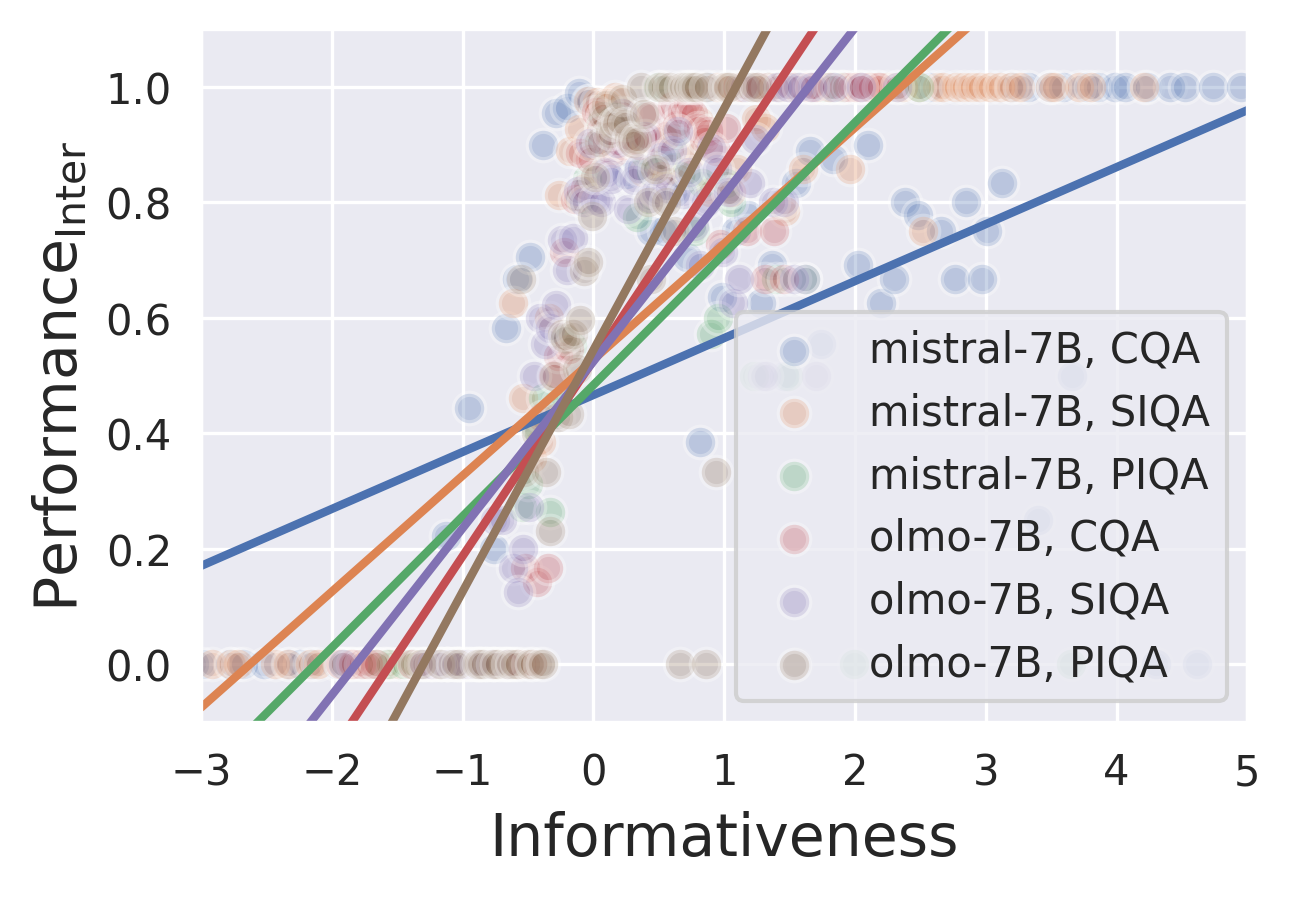}
    \caption{The relationship between Informativeness and Performance$_\text{Inter}$ across six different settings from the stratified correlation analysis (CQA$=$CommonsenseQA, SIQA$=$SocialIQA).}
    \label{fig:informative_performance}
\end{figure}

As shown in Figure~\ref{fig:informative_performance}, the measured Informativeness is positively correlated with Performance$_\text{Inter}$ using CoT. The correlation is not perfect due to the cases where high informativeness still fails to correct predictions where the model is confidently wrong at the beginning.

\subsection{Additional Analyses}
\label{sec:additional_analysis}
To further strengthen the empirical correlation results, we replicate our analysis in two additional settings. We first analyze Mistral-7B's CoT behavior on SocialIQA, which has a similar level of vulnerability to confirmation bias as CommonsenseQA. Second, we evaluate the CoT behavior of OLMo2-7B on CommonsenseQA, using OLMo2-7B as a representative model with weaker internal beliefs (Figure~\ref{fig:comprehensive_entropy_histogram}). 

\subsubsection{Mistral-7B on SocialIQA}

\begin{figure}[!th]
    \centering
    \includegraphics[width=\columnwidth]{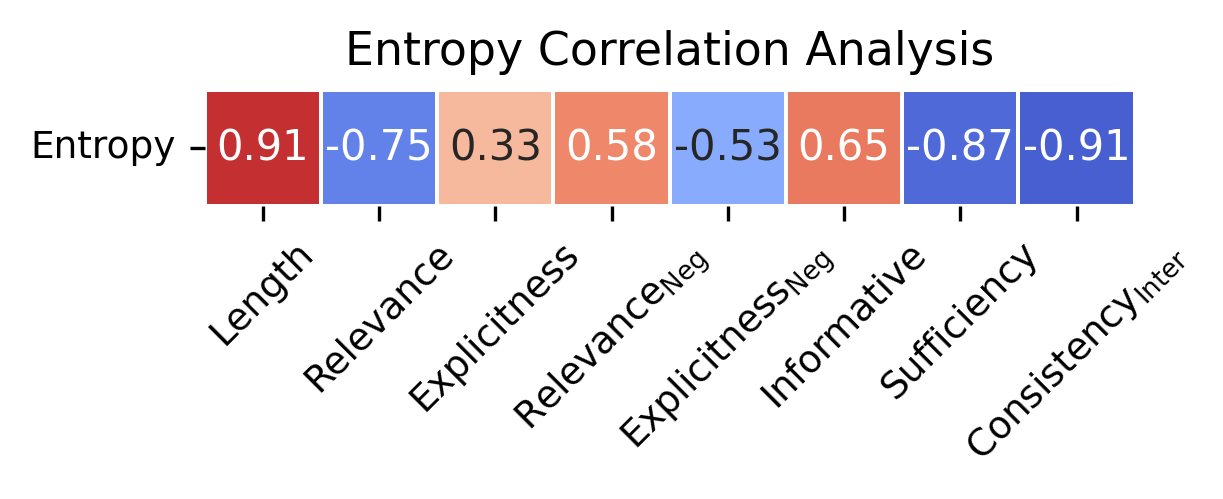}
    \caption{Correlation of Entropy, proxy for strength of model's internal beliefs $B$, with other factors using behaviors of Mistral-7B on SocialIQA.}
    \label{fig:entropy_correlation_SIQA}
\end{figure}

\begin{figure}[!h]
    \centering
    \begin{subfigure}{\columnwidth}
        \centering
        \includegraphics[width=\linewidth]{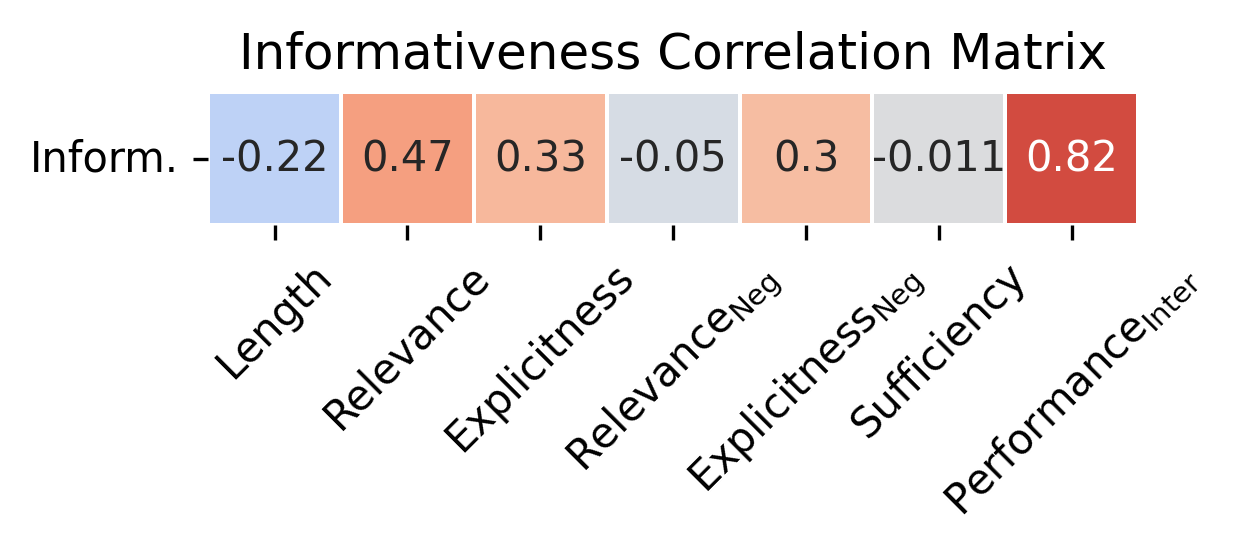}
        \caption{Correlation of Informativeness with other factors.}
        \label{fig:pmi_correlation_SIQA}
    \end{subfigure}
    \hfill
    \begin{subfigure}{\columnwidth}
        \centering
        \includegraphics[width=\linewidth]{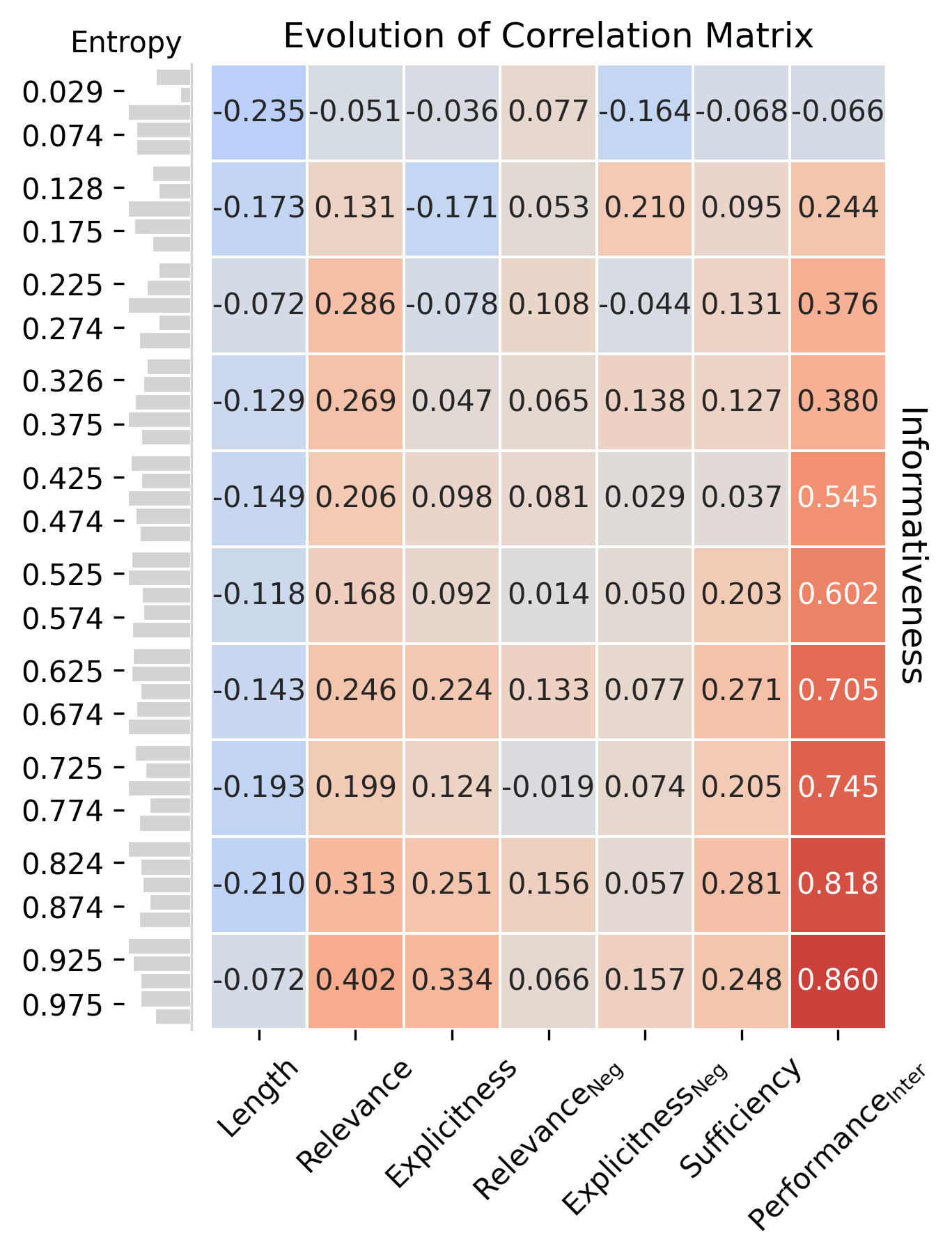}
        \caption{Evolutionary correlation patterns of Informativeness with other factors across different Entropy groups.}
        \label{fig:pmi_correlation_evolve_SIQA}
    \end{subfigure}
    \caption{Correlation analysis of the role of $B$ in the second reasoning stage of $P(A|Q,R,B)$, using behaviors of Mistral-7B on SocialIQA.}
    \label{fig:stage_two_analysis_SIQA}
\end{figure}

\begin{figure*}[t]
    \centering
    \includegraphics[width=\textwidth]{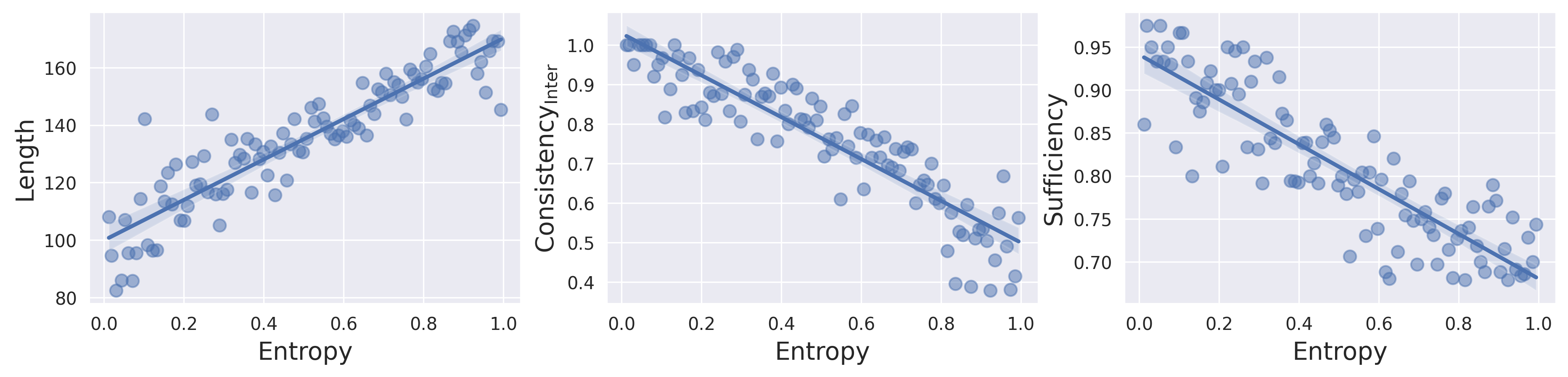}
    \caption{Correlation trends of base entropy (proxy for model's internal beliefs) with CoT Length, Consistency$_\text{Inter}$, and Sufficiency. (Mistral-7B on SocialIQA)}
    \label{fig:example_entropy_correlation_SIQA}
\end{figure*}

We replicate the correlation analysis in the main text and evaluate the CoT behavior of Mistral-7B on SocialIQA. Figure~\ref{fig:entropy_correlation_SIQA} and Figure~\ref{fig:example_entropy_correlation_SIQA} show the stage-one correlation between Entropy (strength of beliefs $B$) and key attributes of rationales generated via $P(R|Q,B)$. Most factors are strongly correlated with Entropy, providing strong evidence of confirmation bias during the first stage of reasoning generation ($Q \to R$). We also include the correlation analysis of stage-two performance in Figure~\ref{fig:stage_two_analysis_SIQA}. Similarly, Figure~\ref{fig:pmi_correlation_evolve_SIQA} demonstrates evolutionary correlation patterns of Relevance, Explicitness, and Sufficiency with Informativeness across different Entropy groups. These results further strengthen the observations discussed in the main text. Even though the exact correlation patterns in Figure~\ref{fig:entropy_correlation_SIQA} and Figure~\ref{fig:stage_two_analysis_SIQA} are slightly different from those in Figure~\ref{fig:entropy_correlation} and Figure~\ref{fig:stage_two_analysis}, this can be attributed to the intrinsic differences in the required reasoning abilities and problem-solving protocols across datasets.

\subsubsection{OLMo2-7B on CommonsenseQA}

\begin{figure}[!h]
    \centering
    \includegraphics[width=\columnwidth]{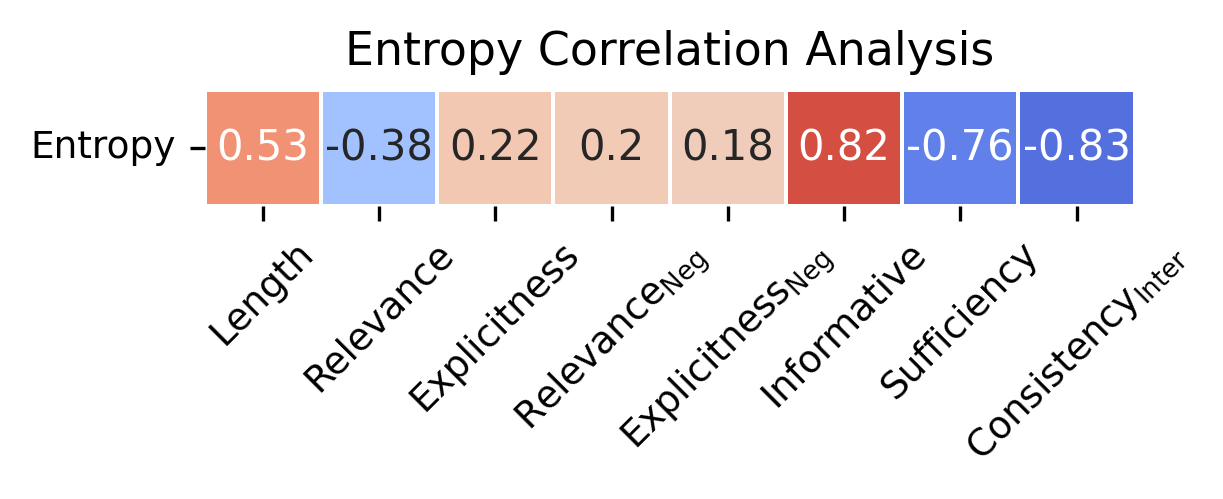}
    \caption{Correlation of Entropy, proxy for strength of model's internal beliefs $B$, with other factors using behaviors of OLMo2-7B on CommonsenseQA.}
    \label{fig:entropy_correlation_OLMo}
\end{figure}

\begin{figure}[!h]
    \centering
    \begin{subfigure}{\columnwidth}
        \centering
        \includegraphics[width=\linewidth]{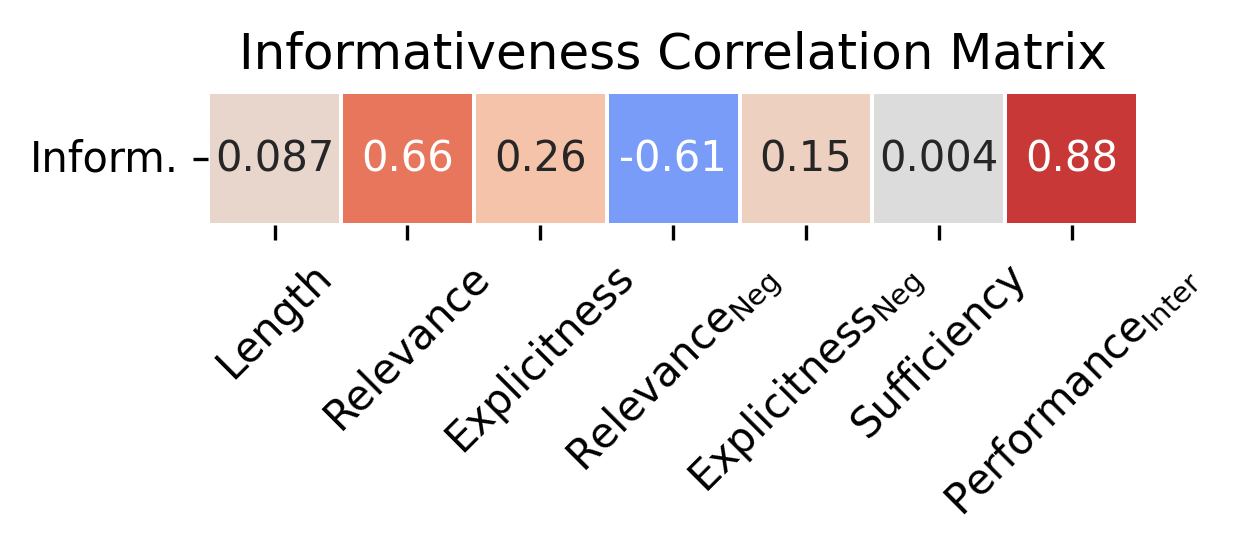}
        \caption{Correlation of Informativeness with other factors.}
        \label{fig:pmi_correlation_OLMo}
    \end{subfigure}
    \hfill
    \begin{subfigure}{\columnwidth}
        \centering
        \includegraphics[width=\linewidth]{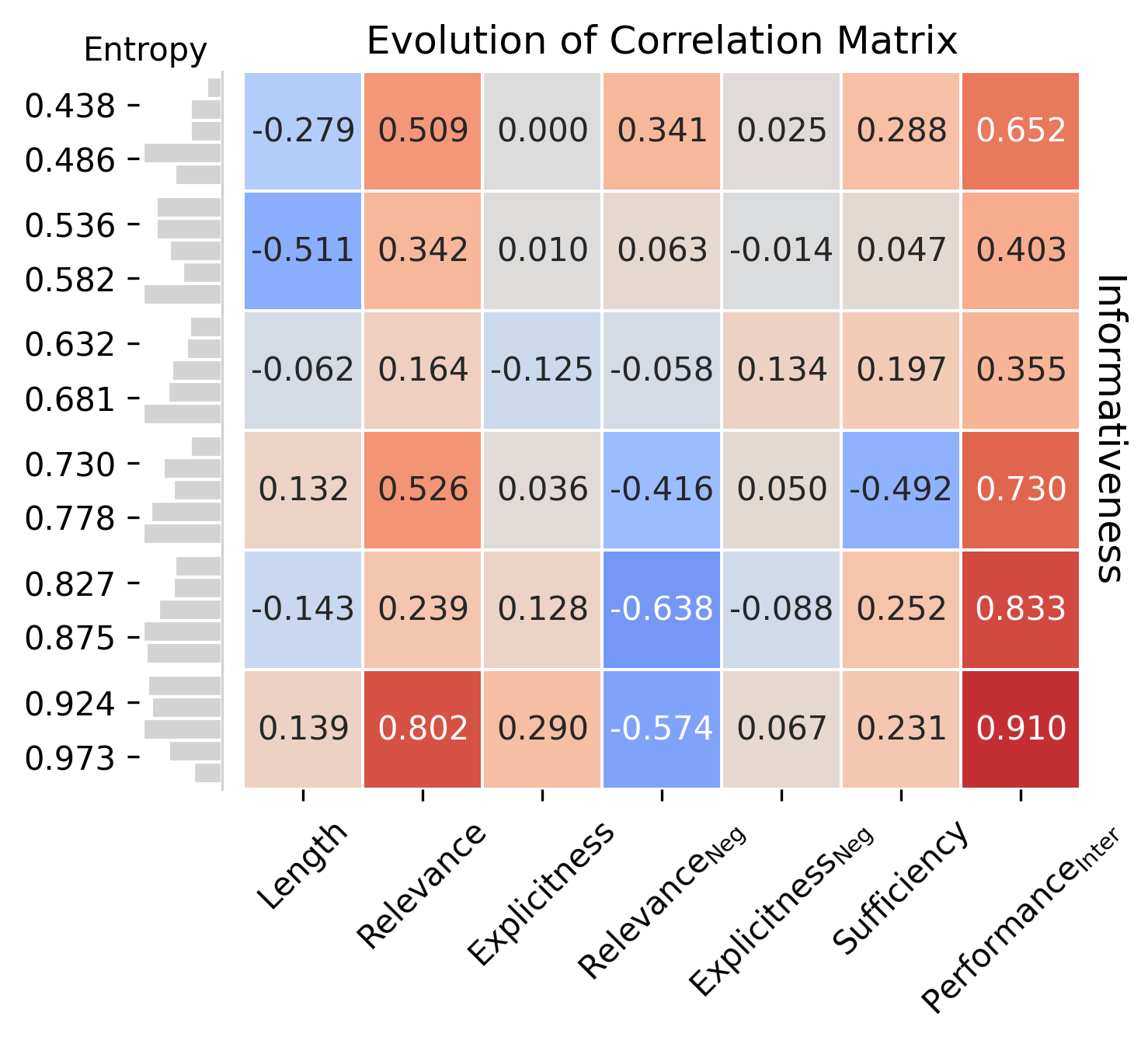}
        \caption{Evolutionary correlation patterns of Informativeness with other factors across different Entropy groups.}
        \label{fig:pmi_correlation_evolve_OLMo}
    \end{subfigure}
    \caption{Correlation analysis of the role of $B$ in the second reasoning stage of $P(A|Q,R,B)$, using behaviors of OLMo2-7B on CommonsenseQA.}
    \label{fig:stage_two_analysis_OLMo}
\end{figure}

\begin{figure*}[t]
    \centering
    \includegraphics[width=\textwidth]{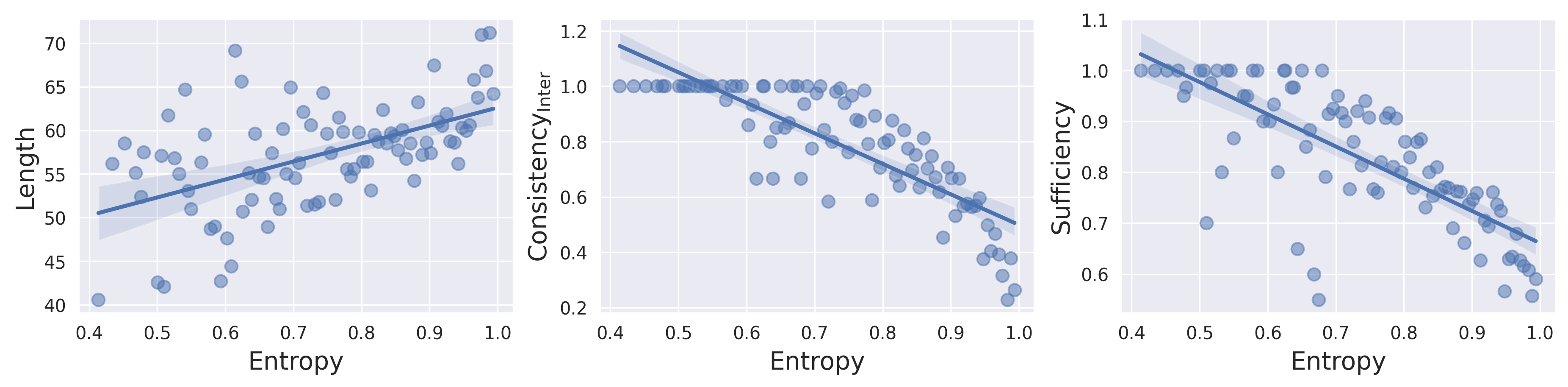}
    \caption{Correlation trends of base entropy (proxy for model's internal beliefs) with CoT Length, Consistency$_\text{Inter}$, and Sufficiency. (OLMo2-7B on CommonsenseQA)}
    \label{fig:example_entropy_correlation_OLMo}
\end{figure*}

We further examine the CoT behavior of OLMo2-7B on CommonsenseQA. Figure~\ref{fig:entropy_correlation_OLMo} and Figure~\ref{fig:example_entropy_correlation_OLMo} show the stage-one correlation between Entropy (strength of beliefs $B$) and key attributes of rationales generated via $P(R|Q,B)$. Even though OLMo2-7B has shown to have weaker beliefs (more high entropy questions) in CommonsenseQA compared to Mistral-7B (Figure~\ref{fig:comprehensive_entropy_histogram}), its Entropy values still correlate substantially with Length, Relevance, Informativeness, Sufficiency, and Consistency$_\text{Inter}$, indicating signs of confirmation bias. We also include the correlation analysis of stage-two performance in Figure~\ref{fig:stage_two_analysis_OLMo}. In contrast to Mistral-7B, OLMo2-7B displays less obvious evolutionary correlation patterns, with only Explicitness and Relevance$_\text{Neg}$ demonstrating clear patterns. This could be attributed to the fact that OLMo2-7B is inherently less prone to confirmation bias. Again, although the exact correlation patterns between Mistral-7B and OLMo2-7B are not the same, it can be explained by differences in the models' problem-solving approaches, which stem from variations in their respective training processes.



\begin{table*}[]
\centering
\renewcommand{\arraystretch}{1.1} 
\begin{tabular}{llccc}
\Xhline{1.5pt}
Dataset               & Model                       & Explicitness & Explicitness$_\text{Neg}>0$ & Performance$_\text{Inter}$ \\ \hline
\multirow{4}{*}{CommonsenseQA}  & \multirow{4}{*}{Mistral-7B} & False        & False                            & 0.821       \\
                      &                             & False        & True                             & 0.783       \\
                      &                             & True         & False                            & 0.963       \\
                      &                             & True         & True                             & 0.965       \\ \hline
\multirow{4}{*}{SocialIQA} & \multirow{4}{*}{Mistral-7B} & False        & False                            & 0.813       \\
                      &                             & False        & True                             & 0.830       \\
                      &                             & True         & False                            & 0.955       \\
                      &                             & True         & True                             & 0.948       \\ \hline
\multirow{4}{*}{CommonsenseQA}  & \multirow{4}{*}{OLMo2-7B}   & False        & False                            & 0.873       \\
                      &                             & False        & True                             & 0.842       \\
                      &                             & True         & False                            & 0.977       \\
                      &                             & True         & True                             & 0.953       \\ \Xhline{1.5pt}
\end{tabular}
\caption{Average reasoning-following performance ($QR \to A$), Performance$_\text{Inter}$, with respect to rationales' Explicitness and Explicitness$_\text{Neg}$ levels.}
\label{tab:explicit}
\end{table*}

\clearpage

\begin{table*}[]
\centering
\renewcommand{\arraystretch}{1.3} 
\small
\begin{tabular}{p{0.16\textwidth}p{0.78\textwidth}}
\Xhline{1.5pt}
\textbf{Attribute} & \textbf{Description} \\ \hline
Length & 
We mainly measure the token-level length of the reasoning. \newline \textit{Formulation:} $N$
\\ \hline
Relevance & 
The query relevance score \cite{towardsunderstandcot} measures whether the reasoning step merely explains the question itself or reasons towards the connection between the question and the answer $A_\text{inter}$. In this work, query relevance is first computed at the step-level using textual entailment between each reasoning step $R_i$ and a predefined explanation hypothesis in the form of \textit{"the sentence is talking about ..."}. The step-level entailment probabilities are then averaged to obtain the overall rationale-level relevance score.
\newline 
\textit{Formulation:} $\frac{1}{T}\sum_i^T R_i \models  \text{explain} (A_\text{inter})$ 
\\ \hline 
Relevance$_\text{Neg}$ & 
The \underline{Neg}ative relevance score measures whether the reasoning step explains why alterative options other than $A_\text{inter}$ are wrong. To compute this, we first measure the entailment probability between each reasoning step and the alternative answer choices. The final rationale-level score is obtained by averaging these entailment probabilities across both the answer choices and the reasoning steps.
\newline
\textit{Formulation:} $\frac{1}{M-1} \frac{1}{T} \sum_{A_j \ne A_\text{inter}} \sum_i^T R_i \models \text{explain} (A_{j})$
\\ \hline
Explicitness &
It is common for models to state explicit conclusion (e.g., \textit{"... is the most appropriate answer."}) in the middle of step-by-step reasoning. We observe that it has a strong influence on subsequent reasoning and the final prediction (Appendix~\ref{sec:explicitness}). Similar to relevance, explicitness is first measured at step-level using textual entailment between $R_i$ and the conclusion hypothesis of $A_\text{inter}$ in the form of \textit{"the answer is ..."}, and aggregated into the rationale-level explicitness score. Note that this score is a more extreme form of relevance score.
\newline
\textit{Formulation:} $\frac{1}{T}\sum_i^T R_i \models \text{conclude} (A_\text{inter})$
\\ \hline
Explicitness$_\text{Neg}$ &
The main idea of this score is similar to the explicitness score but focuses on explicit rejection (e.g., \textit{"... is impossible."}). Again, we first measure textual entailment between each reasoning step $R_i$ and the rejection of answer choices in the form of \textit{"the answer is not ..."}. The final rationale-level rejection score is then obtained by averaging the entailment probabilities across both the answer choices and reasoning steps.
\newline
\textit{Formulation:} $\frac{1}{M-1} \frac{1}{T} \sum_{A_j \ne A_\text{inter}} \sum_i^T R_i \models \text{reject} (A_{j})$
\\ \hline
Informativeness &
We leverage the concept of point-wise mutual information (PMI), following the work in \cite{Bosselut2020DynamicNK, holtzman-etal-2021-surface}, to quantify how much additional information the reasoning process provides in supporting the decision of answer $A_\text{inter}$. A highly PMI value indicates that the CoT is more likely to conclude with $A_\text{inter}$. This metric is highly correlated with Performance$_\text{Inter}$ (Appendix~\ref{sec:informativeness}).
\newline
\textit{Formulation:} $\log P(A_\text{inter}|Q, R) / P(A_\text{inter}|Q)$
\\ \hline
Sufficiency &
The reasoning sufficiency is evaluated by predicting the answer using only the rationale ($R \to A$). We argue that, if the reasoning is sufficient enough, it should yield the same answer as the full reasoning $QR \to A$, even without accessing the question.
\newline
\textit{Formulation:} $\mathbb{I}\left(\text{argmax}_i P(A_i|R) = \text{argmax}_iP(A_i|Q, R)\right)$
\\ \hline
Consistency$_\text{Inter}$ &
Intermediate (Inter) reasoning consistency examines whether the answer choice supported by the rationale, $A_\text{inter}$, aligns with the model’s initial prediction from $Q \to A$. In other words, it evaluates whether the rationale reinforces the model's original belief or causes a shift in its answer choice.
\newline
\textit{Formulation:} $\mathbb{I}\left(A_\text{inter} = \text{argmax}_iP(A_i|Q)\right)$
\\ \hline
Performance$_\text{Inter}$ &
This metric measures whether the predicted answer choice, given the rationale, matches the answer $A_\text{inter}$ supported by the rationale. In other words, it solely assesses the performance of the stage $QR \to A$.
\newline
\textit{Formulation:} $\mathbb{I}\left(\text{argmax}_iP(A_i|Q,R) = A_\text{inter}\right)$
\\ \hline
Performance$_\text{E2E}$* & 
This is the conventional performance metric that measure whether the predicted answer choice matches the ground truth label. 
\newline
\textit{Formulation:} $\mathbb{I}\left(\text{argmax}_iP(A_i|Q,R) = A^*\right)$
\\ \Xhline{1.5pt}
\end{tabular}
\caption{Evaluation metrics for rationale. The asterisk (*) denotes that the metric requires access to the annotated ground truth label.}
\label{tab:metric}
\end{table*}

\end{document}